# Heavy Rain Face Image Restoration: Integrating Physical Degradation Model and Facial Component Guided Adversarial Learning

[*]Chang-Hwan Son and Da-Hee Jeong

Department of Software Science & Engineering, Kunsan National University

558 Daehak-ro, Gunsan-si 54150, Republic of Korea

[*]Corresponding Author

Phone Number: 82-63-469-8915; Fax Number: 82-63-469-7432

E-MAIL: changhwan76.son@gmail.com; cson@kunsan.ac.kr

**Abstract**

With the recent increase in intelligent CCTVs for visual surveillance, a new image degradation that integrates resolution conversion and synthetic rain models is required. For example, in heavy rain, face images captured by CCTV from a distance have significant deterioration in both visibility and resolution. Unlike traditional image degradation models (IDM), such as rain removal and superresolution, this study addresses a new IDM referred to as a *scale-aware heavy rain model* and proposes a method for restoring high-resolution face images (HR-FIs) from **l**ow-**r**esolution **h**eavy **r**ain **f**ace **i**mages (LRHR-FI). To this end, a 2-stage network is presented. The first stage generates low-resolution face images (LR-FIs), from which heavy rain has been removed from the LRHR-FIs to improve visibility. To realize this, an interpretable IDM-based network is constructed to predict physical parameters, such as rain streaks, transmission maps, and atmospheric light. In addition, the image reconstruction loss is evaluated to enhance the estimates of the physical parameters. For the second stage, which aims to reconstruct the HR-FIs from the LR-FIs outputted in the first stage, facial component guided adversarial learning (FCGAL) is applied to boost facial structure expressions. To focus on informative facial features and reinforce the authenticity of facial components, such as the eyes and nose, a face-parsing-guided generator and facial local discriminators are designed for FCGAL. The experimental results verify that the proposed approach based on physical-based network design and FCGAL can remove heavy rain and increase the resolution and visibility simultaneously. Moreover, the proposed heavy-rain face image restoration outperforms state-of-the-art models of heavy rain removal, image-to-image translation, and superresolution.

**1. Introduction**

Captured images sometimes have unwanted degradation caused by noise, blurring, downsampling, rain streaks, etc.. To restore a high-quality image from such a degraded image, understanding how the image degradation process is mathematically formulated is necessary. Image restoration refers to the reversal of image degradation models (IDMs) [1]. Traditional image restoration involves the following IDM:

$$(\mathbf{H} \otimes \mathbf{K}) \downarrow_s + \mathbf{n} = \mathbf{J} \quad (1)$$

where $\mathbf{H}$ and $\mathbf{J}$ are the high-quality original image and the degraded measured image, respectively, $\mathbf{K}$ the motion filter, $\otimes$ the convolution operator, $\downarrow_s$ the downsampling operator with a scale factor of $s$, and $\mathbf{n}$ is the noise. Eq. (1) is the IDM for a **s**upe**r**resolution (SR) with a Gaussian filter $\mathbf{K}$, but it can also be used for denoising and deblurring. If the motion kernel is set with a delta function, and the downsampling operation is omitted, Eq. (1) becomes the IDM for denoising. If only the downsampling operation is removed, then Eq. (1) represents the IDM for deblurring [2].

**1.1 Necessity of a new scale-aware heavy rain model**

Recently, intelligent CCTVs have been actively developed for visual surveillance purposes, such as pedestrian detection, object counting, and abnormal behavior detection. Because CCTVs are installed outdoors, they are placed in adverse weather conditions such as heavy rain, dense fog, and heavy snow, which interfere with the acquisition of high-quality images [3-6]. Fig. 1 shows an example of **l**ow-**r**esolution **h**eavy **r**ain **f**ace **i**mages (LRHR-FI) with severe degradation in visibility and resolution, which were taken at a distance from the cameras. Recognizing facial images is difficult because of heavy rains.

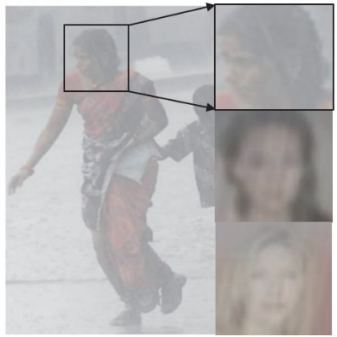

Fig. 1. Example of LRHR-FIs.

In addition to the traditional IDMs in Eq. (1), the rain streak and heavy rain models are introduced for rain removal. Their mathematical IDMs are represented by $\mathbf{I} = \mathbf{J} + \sum_i^m \mathbf{S_i}$ [7] and $\mathbf{I} = \mathbf{T} \odot (\mathbf{J} + \sum_{i=1}^m \mathbf{S_i}) + (\mathbf{1} - \mathbf{T}) \odot \mathbf{A}$ [8], respectively. Here, $\mathbf{T}$ is the transmission introduced by the scattering process of tiny water particles, $\mathbf{A}$ the atmospheric light, $\mathbf{S_i}$ the rain layer containing rain streaks, $\mathbf{1}$ a matrix of ones, $\odot$ element-wise multiplication, and $\mathbf{I}$ and $\mathbf{J}$ the degraded rain image and original image, respectively. Compared to the rain streak model, the heavy rain model reflects the veiling effect, which makes the scene look misty and reduces the visibility. The veiling effect is the result of rain-streak accumulation in the sight.

However, the rain synthesis models and classical IDMs introduced thus far are now inappropriate for super-resolving LRHR-FIs, as shown in Fig. 1. Therefore, a new IDM is required to predict the SR **f**ace **i**mage (SR-FI) from the captured LRHR-FI. To the best of our knowledge, this study is the first to introduce a new IDM, referred to as a *scale-aware heavy rain model,* and propose a novel approach for solving the inverse problem of the following IDM:

$$\mathbf{I} = \mathbf{T} \odot \left( (\mathbf{H} \otimes \mathbf{K}) \downarrow_s + \sum_{i=1}^{m} \mathbf{S_i} \right) + (\mathbf{1} - \mathbf{T}) \odot \mathbf{A} \tag{2}$$

where **I** and **H** are the LRHR-FI and original **h**igh-**r**esolution **f**ace **i**mage (HR-FI), respectively. Depending on the context, the HR refers to **h**eavy **r**ain or **h**igh **r**esolution. In Eq. (2), note that the scale-aware heavy rain model includes downsampling and blurring operations, i.e., $(\mathbf{H} \otimes \mathbf{K}) \downarrow_s$. This differs from the existing heavy-rain model [8]. Therefore, the scale-aware heavy rain model integrates low-resolution conversion and a synthetic heavy rain model.

**1.2 Proposed approach**

The inverse problem in Eq. (2), is challenging. This is because Eq. (2) has high-dimensional physical parameters **T, S** and **A,** to be estimated, and the reverse of convolution and downsampling must be performed. In addition, the captured LRHR-FIs exhibit severe deterioration in visibility and resolution. Thus, the inverse problem is highly ill-posed. However, Eq. (2) can be decomposed into two terms as follows:

$$\mathbf{I} = \mathbf{T} \odot \left( \mathbf{J} + \sum_{i=1}^{m} \mathbf{S_i} \right) + (\mathbf{1} - \mathbf{T}) \odot \mathbf{A} \tag{3}$$

$$\mathbf{J} = (\mathbf{H} \otimes \mathbf{K}) \downarrow_s \tag{4}$$

These equations indicate that the reverse of Eq. (2) represents the integration of heavy-rain removal and SR. Therefore, this study proposes a unified framework for super-resolving LRHR-FIs. Fig. 2 shows the proposed unified framework for joint heavy-rain removal and SR. To address the inverse problems in Eqs. (3) and (4), an interpretable IDM-based network and facial component guided adversarial learning (FCGAL) are employed.

For heavy-rain removal, precisely predicting the physical parameters **T, S,** and **A** is important in Eq. (3). To reflect this, an interpretable IDM-based network is designed for physics-based rain removal. In addition to the mean squared error (MSE), image reconstruction loss, which can measure the difference between the captured LRHR-FI and predicted LRHR-FI, is used as a regularization term to enhance the estimates of the physical parameters. FCGAL aims to boost facial structure expressions. The details of facial components, such as the eyes, lips, and nose, are crucially important for face SR. To increase the discriminative power of facial features and reinforce the authenticity of facial components, a face-parsing-guided generator and local discriminators are added to the conventional generative adversarial network(GAN). The proposed unified network learns in an end-to-end manner for the reverse of the scale-aware heavy rain model.

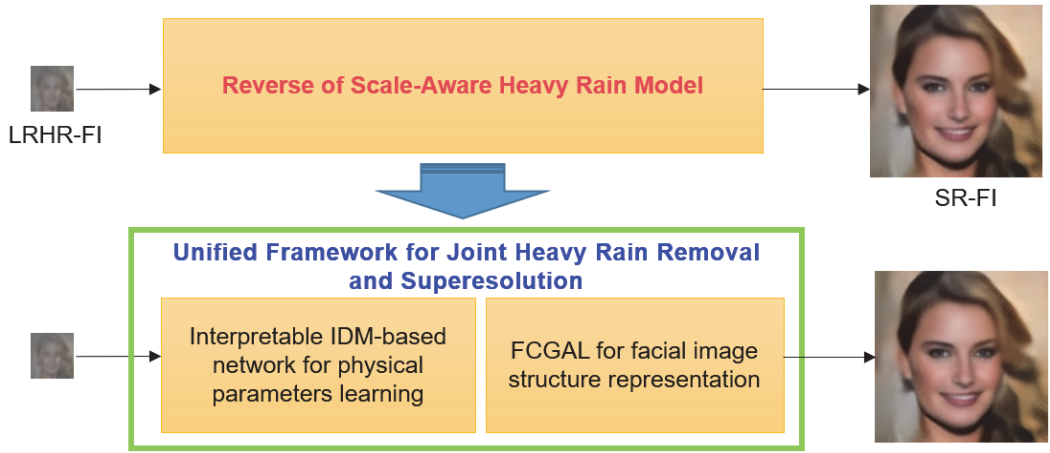

Fig. 2. Proposed unified framework for the reverse of the scale-aware heavy rain model.

**1.3 Contributions**

- To the best of our knowledge, this study is the first to introduce a new IDM, referred to as the scale-aware heavy rain model. With the rapid increase in intelligent CCTVs, face images can be captured with severe degradation in resolution and visibility, complicating face recognition. To address this issue, the proposed scale-aware heavy rain model that integrates low-resolution conversion and a synthetic rain model is required for accurate face image restoration.

- This study proposes a unified framework for joint heavy rain removal and SR by employing an interpretable IDM-based network and FCGAL. For the inverse of the scale-aware heavy-rain model, an interpretable IDM-based network is designed for physics-based heavy-rain removal. In addition, for improved facial structure expressions, FCGAL is proposed for enabling the facial attention mechanism and learn local discriminators for facial authenticity examination.

- This study provides new training and test datasets for super-resolving LRHR-FIs with low resolutions and visibility. To generate synthetic LRHR-FIs, CelebA-HQ, which includes clean facial images, is used. The source code for generating LRHR-FIs, according to Eq. (2), is available to the public for research purposes. This dataset can be used as a reference dataset, and the evaluation scores can be used for performance comparison.

## 2. Related Work

This study addresses the inverse problem of the new scale-aware heavy rain model for super-resolving LRHR-FIs and proposes a unified framework for joint heavy-rain removal and SR based on the interpretable IDM-based network design and FCGAL. Therefore, this section reviews related work on rain removal and SR.

### 2.1. Rain removal

Single-image rain removal methods [9-16] can be categorized into model-based and data-driven methods. Model-based approaches employ optimization frameworks consisting of a data fidelity term and a prior term. The data-fidelity term measures the accuracy of rain synthesis models such as additive [9] and nonlinear composite models [8]. The prior term models handcrafted priors regarding rain shape and direction.

$$\hat{\mathbf{J}} = \arg\min_{\mathbf{J}} \|\mathbf{I} - \mathbf{S} - \mathbf{J}\|_2^2 + \Omega(\mathbf{S}) + \Upsilon(\mathbf{J}) \tag{5}$$

where the first term represents the data-fidelity term. In Eq. (5), it is assumed that an additive rain streak model is used. For heavy rain removal, the rain streak model should be replaced with a heavy rain model. The second and third terms are prior terms. The core of solving Eq. (5) is prior modeling. To date, various types of models have been designed for improved description. Sparse coding [6,9,10], Gaussian mixture models (GMM) [11] and directional prior modelling [12] are the most popular approaches for prior modeling before the development of deep learning methods.

Data-driven approaches use a large amount of observed data and automatically extract rich hierarchical features through a layer-by-layer transformation. Data-driven approaches are deep learning techniques. Unlike the prior models mentioned above, deep learning can provide abundant features without requiring iterations for optimization; that is, it can directly predict the de-rained images through the pretrained network.

$$\hat{\mathbf{J}} = f_{\boldsymbol{\theta}}(\mathbf{I}) \tag{6}$$

Here, $f_{\boldsymbol{\theta}}$ indicates a network with a learnable parameter $\boldsymbol{\theta}$ for rain removal. $f_{\boldsymbol{\theta}}$ takes a rain image $\mathbf{I}$ as input and outputs a derained image $\hat{\mathbf{J}}$.

Since the introduction of the detailed network [13], various architectures such as density-aware [14], joint rain detection and removal [7], scale-aware [15,16], and progressive networks [17], have been developed. However, these deep learning models target rain streaks and heavy-rain images. Therefore, these are expected to be unsuitable for the scale-aware heavy rain model focused on in this study. That is, low-resolution and visibility problems cannot be addressed jointly.

## 2.2. Superresolution

SR technology started with a classical approach based on adaptive filtering and interpolation [18] and has transitioned to machine learning approaches based on sparse representation [19], principal component analysis [20], and image priors [21-23]. Similar to rain removal methods, SR approaches can be divided into model-based and data-driven approaches.

$$\hat{\mathbf{H}} = \arg\min_{\mathbf{H}} \|(\mathbf{H} \otimes \mathbf{K}) \downarrow_s - \mathbf{J}\|_2^2 + \Upsilon(\mathbf{H}) \tag{7}$$

Here, the first term is the data-fidelity term used to model the IDM for SR and the second is the prior term used to characterize HR images, for example, sparsity [19], gradient shape [21], and patch redundancy [23].

Recently, deep learning has become a mainstream approach. Similar to deep-learning-based rain removal, layer-by-layer transformation is performed to extract rich features and reconstruct SR images.

$$\hat{\mathbf{H}} = f_{\boldsymbol{\theta}}(\mathbf{J}) \tag{8}$$

In this equation, the SR network $f_{\boldsymbol{\theta}}$ takes LR image **J** and outputs SR image $\hat{\mathbf{H}}$. This means that an upsampling layer is required in the network to match the input and output image sizes. Bicubic and transposed convolution layers are primarily used for upsampling.

Deep-learning-based SR methods can be classified as MSE-oriented and perceptual-driven [24]. MSE-oriented approaches focus on minimizing pixel-wise distances between the SR and HR images. Such an optimization objective induces deep learning to produce SR images, which may be a statistical average of the possible HR solutions. Consequently, blurry images with a high peak signal-to-noise ratio (PSNR) can be generated [24]. VDSR[25] and RCAN [26] are representative MSE-oriented SR models. In contrast, perceptual-driven approaches aim to recover photo-realistic images within the GAN framework. In this framework, a new perceptual loss function is used to measure the distance between the pretrained features of the SR and HR images [24]. Popular models include SRGAN [24] and ESRGAN [27]. However, perceptual-driven SR models tend to produce structural distortions, along with fine textures. To address this issue, the SPSR [28] model with gradient guidance is proposed.

For face SR, side information [29-31] such as facial landmarks and masks, have been exploited to improve facial details. Spatial transformation networks (STN) [32] can be considered to be agnostic for face poses and deformation. More recently, blind face restoration [33] has been actively studied to handle complex and unknown degradation. However, these face SR methods are not designed for super-resolving LRHR-FIs. To remove rain streaks and increase the resolution and visibility, a unified framework for jointly learning heavy rain removal and SR is required. In addition, employing local discriminators and face-parsing-guided generators that can extract more informative features in the facial areas and reinforce the authenticity of the facial components is better.

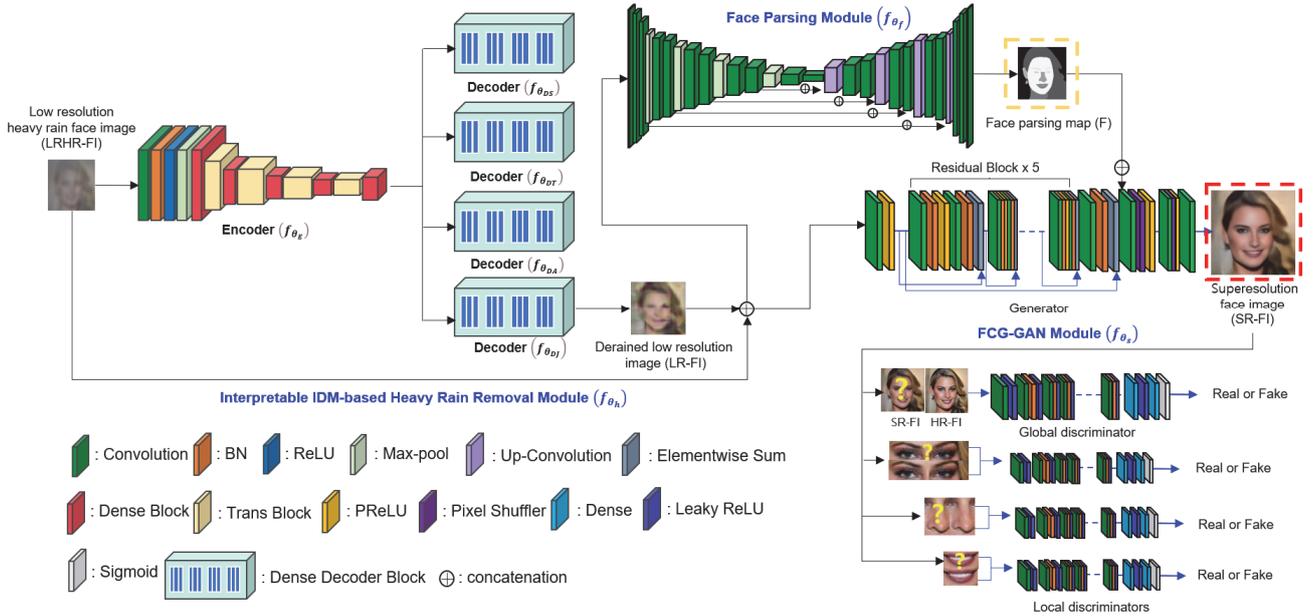

Fig. 3. Proposed deep learning network for heavy rain face image restoration.

## 3. Proposed Heavy Rain Face Image Restoration

This study is interested in a unified deep learning framework for the inverse of a new scale-aware heavy rain model that integrates low-resolution conversion (e.g., downsampling) and a rain synthesis model. The inverse process can be considered as a joint heavy-rain removal and SR. Although the rain removal and SR models discussed in the previous section are state-of-the-art (SOTA), they are unsuitable for joint heavy-rain removal and SR. Heavy-rain removal models only consider rain streaks and accumulation; thus, HR conversion is unachievable. Similarly, because SR models are concerned only with HR image reconstruction, visibility cannot be addressed because of heavy rain. To this end, a new unified deep learning framework for joint heavy rain removal and SR is proposed.

Fig. 3 shows the architecture of the proposed network for heavy rain face image restoration that incorporates interpretable IDM-based heavy-rain removal and FCGAL for SR. As shown in the input LRHR-FI, the face image exhibits severe degradation in visibility and resolution. Thus, face recognition is difficult, and the inverse problem of the proposed scale-aware heavy-rain model is challenging. The proposed network comprises three modules: IDM-based heavy-rain removal, face parsing, and facial component-guided GAN (FCG-GAN). The IDM-based heavy rain removal module is used for de-rained LR image reconstruction by removing rain streaks and accumulation. The face parsing module aims to find facial regions and provide output that guides facial attention modeling. To this end, its output map is incorporated into the generator of the proposed FCG-GAN. In addition, in FCG-GAN, local facial discriminators are designed to reinforce the authenticity of facial components, such as the eyes and nose. Ultimately the FCN-GAN aims to boost facial structure expression for better SR image reconstruction. The details of the three modules are provided below. The notations are summarized in Table 1 before the details of the proposed network are introduced.

Table 1. Notation

| Notation | Definition |
|---|---|
| I | **L**ow-**r**esolution **h**eavy **r**ain **f**ace **i**mages (LRHR-FI) |
| J | **L**ow-**r**esolution **f**ace **i**mages (LR-FI) |
| H | **H**ow-**r**esolution **f**ace **i**mages (HR-FI) |
| F | Parsed **l**ow-**r**esolution **f**ace **i**mage (Parsed LR-FI) |
| S | Rain layer |
| T | Transmission map |
| A | Atmospheric light map |

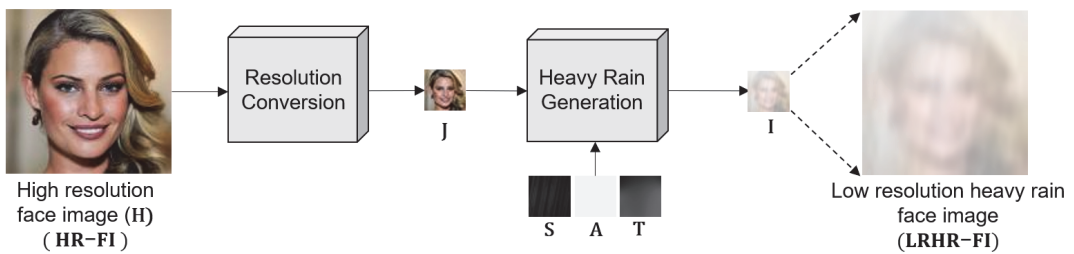

Fig. 4. Synthetic LRHR-FI generation based on the proposed scale-aware heavy rain model.

## 3.1 Synthetic image generation

Although introduced in Subsection 1.3, to reiterate, the proposed scale-aware heavy rain model has the following mathematical form:

$$\mathbf{I} = \mathbf{T} \odot \left((\mathbf{H} \otimes \mathbf{K}) \downarrow_s + \sum_{i=1}^{m} \mathbf{S_i}\right) + (\mathbf{1} - \mathbf{T}) \odot \mathbf{A} \qquad (9)$$

where **I** and **H** are the LRHR-FI and HR-FI, respectively; **T**, **A**, and **S** the transmission map, atmospheric light, and rain layer, respectively; **1** a matrix of ones; $\odot$ element-wise multiplication; **K** the Gaussian filter, $\otimes$ the convolution operator; and $\downarrow_s$ is the downsampling operator with scale factor $s$.

The proposed scale-aware heavy rain model consists of two steps: resolution conversion and heavy-rain generation, as illustrated in Fig. 4. For LRHR-FI (**I**) generation, the original HR-FIs (**H**) are prepared. In this study, the commonly used CelebA-HQ[33] dataset is selected for HR-FIs. According to Eq. (9), HR-FI (**H**) is first convolved with a Gaussian filter and downsampled with scale factor $s$ to generate LR-FI, $\mathbf{J} = (\mathbf{H} \otimes \mathbf{K}) \downarrow_s$. Next, the rain layer (**S**) including rain streaks is added to the LR-FI, and then blended with the atmospheric light **A** at the ratio of transmission map **T**. Specifically, bicubic interpolation is used to implement $(\mathbf{H} \otimes \mathbf{K}) \downarrow_s$ and $\mathbf{S}_i$ is synthesized by generating the Gaussian noise and applying motion filters. Atmospheric light map ***A*** is filled with the same bright pixel values, and ***T*** is derived from the depth map predicted with the haze model, in which the scene radiance is exponentially diminished with depth [3,4]. A random number generator is used to determine the noise level, direction, length of the motion filter, and pixel value of the atmospheric light.

Fig. 5 shows an example of the synthesized FI according to the proposed scale-aware heavy rain model. In Fig. 5, the first and second images are the original HR-FI (**H**) and its counterpart, LR-FI (**J**), respectively. The image sizes of the HR-FI and LR-FI are 128 × 128 and 32 × 32, respectively. Bicubic interpolation is used to generate the LR-FI from the given HR-FI. Because the size of the LR-FI is too small, the LR-FI is resized to the same size as that of the original HR-FI for visualization. The third image is a rain streak image synthesized using $\mathbf{J} + \sum_{i=1}^{m} \mathbf{S_i}$, where $m$ is set to 1, which means that one rain streak is added. Compared with LR-FI, the rain streak image contains rain patterns. The fourth is the LRHR-FI (**I**) generated using **T**⊙**J** + (**1** − **T**)⊙**A,** where visibility is significantly reduced, because of the blending operation that simulates the veiling effect. The LRHR-FI exhibits a significant degradation in both visibility and resolution. Therefore, reconstructing the HR-FI is challenging from the given LRHR-FI. The last three images are the rain layer (**S**), atmospheric light (**A**) and transmission (**T**). For reference, depth estimation is meaningful for natural scenes; however, for face images, the depth is assumed to be almost the same.

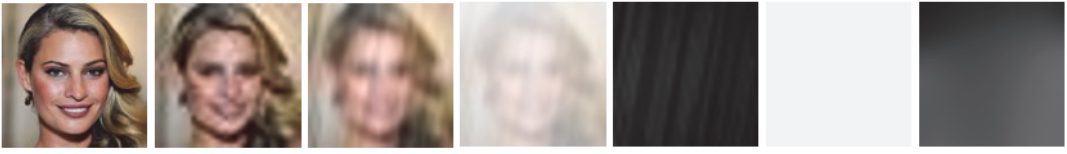

Fig. 5. Examples of synthetic images; HR-FI, LR-FI, rain streaked image, LRHR-FI, rain layer (**S**), atmospheric light (**A**), and transmission (**T**) (left to right).

**3.2 Inverse problem of the proposed scale-aware heavy rain model**

This study intends to solve a new inverse problem for the proposed scale-aware heavy rain model, as shown in Eq. (5), that is, reconstructing the HR-FI from the given LRHR-FI.

$$\hat{\mathbf{H}} = f_{\boldsymbol{\theta}}(\mathbf{I}) \tag{10}$$

where $f_{\boldsymbol{\theta}}$ indicates the proposed network with learnable parameter $\boldsymbol{\theta}$. $f_{\boldsymbol{\theta}}$ takes LRHR-FI (**I**) as input and outputs the SR-FI ($\hat{\mathbf{H}}$). To effectively super-resolve the LRHR-FI, the proposed network is decomposed into three modules, as follows:

$$f_{\boldsymbol{\theta}} = \{f_{\boldsymbol{\theta}_h}, f_{\boldsymbol{\theta}_f}, f_{\boldsymbol{\theta}_s}\} \tag{11}$$

where $f_{\theta_h}$, $f_{\theta_f}$, and $f_{\theta_s}$ denote the three modules: interpretable IDM-based heavy-rain removal, face parsing, and FCN-GAN, respectively.

**3.3 Interpretable IDM-based heavy rain removal module**

In Fig. 3, the IDM-based heavy rain removal module aims to remove rain streaks and accumulation, thereby generating de-rained LR-FIs with improved visibility.

$$\hat{\mathbf{J}} = f_{\theta_h}(\mathbf{I}) \tag{12}$$

where $f_{\theta_h}$ is the heavy rain removal module, that takes the LRHR-FI (**I**) as input and outputs the de-rained LR-FI ($\hat{\mathbf{J}}$). Note that the resolution of the de-rained LR-FI is low. To construct the heavy-rain removal module, this study adopted a physics-based approach to interpret the module and understand the IDM. Although the physical parameters are repeatedly updated based on an optimization algorithm (i.e., half-quadratic splitting [35]), the heavy-rain removal module can be viewed as a simple and interpretable network because it includes the IDM for heavy rain.

In Fig. 3, the proposed IDM-based heavy-rain removal module consists of one encoder and four decoders. The encoder extracts visual features from the LRHR-FI (**I**). During encoding, a high-dimensional image is mapped to low-dimensional feature vectors. Pretrained models such as VGG [36], ResNet [37], and DenseNet [38] have become the basis of encoders. In this study, DenseNet is chosen as the encoder and decoders. The details of the architecture are provided in [38]. Three of the four decoders are used to predict the physical parameters of the rain layer (**S**), transmission map (**T**), and atmospheric light (**A**), and the fourth encoder estimates the derived LR-FIs ($\hat{\mathbf{J}}$).

$$\hat{\mathbf{S}} = f_{\theta_{DS}}\big(f_{\theta_E}(\mathbf{I})\big) \tag{13}$$

$$\hat{T} = f_{\theta_{DT}}(f_{\theta_E}(I)) \tag{14}$$

$$\hat{A} = f_{\theta_{DA}}(f_{\theta_E}(I)) \tag{15}$$

$$\hat{J} = f_{\theta_{DJ}}(f_{\theta_E}(I)) \tag{16}$$

where $f_{\theta_E}$ is the learnable function for the encoder, and $f_{\theta_{DS}}$, $f_{\theta_{DT}}$, $f_{\theta_{DA}}$, and $f_{\theta_{DJ}}$ the learnable functions for the four decoders. To train the $f_{\theta_E}$, $f_{\theta_{DS}}$, $f_{\theta_{DT}}$, $f_{\theta_{DA}}$, and $f_{\theta_{DJ}}$, the following loss function is minimized:

$$L_{R,T} = L_R + \omega_1 L_{vgg} \tag{17}$$

$$L_R = f_{MSE}(\hat{J}, J) + f_{MSE}(\hat{I}, I) \tag{18}$$

$$\hat{I} = \hat{T} \odot (J + \sum_{i=1}^{m} \hat{S}_i) + (1 - \hat{T}) \odot \hat{A} \tag{19}$$

$$L_{vgg} = vgg(\hat{J}, J) + vgg(\hat{I}, I) = \sum_{i=1}^{3} \|g_i(\hat{J}) - g_i(J)\|_2^2 + \sum_{i=1}^{3} \|g_i(\hat{I}) - g_i(I)\|_2^2 \tag{20}$$

The total loss $L_{R,T}$ is comprises $L_R$ and $L_{vgg}$. In Eq. (18), $f_{MSE}$ indicates the function used to calculate the MSE. The first term of $L_R$ measures the reconstruction error between the de-rained LR-FI and original LR-FI, and the second term is the regularization term referred to as the image reconstruction loss that measures the pixel distance between the original LRHR-FI and restored LRHR-FI using Eq. (19). The predicted physical parameters $\hat{T}$, $\hat{A}$, and $\hat{S}_i$ are used for LRHR-FI reconstruction. By adding the second term, the estimates of physical parameters can be enhanced. $L_{vgg}$ is the total perceptual loss, and $vgg$ calculates the perceptual loss to determine the dissimilarity between the high-level features of two input images. In this study, a pretrained VGG model extracts high-level features. In Eq. (20), $g_i$ represents the operation that extracts high-level features from the pretrained VGG model at the $i$th layer. ReLU1_1, ReLU2_2, and ReLU3_3 are selected as layers. $\omega_1$ is the weight value used to balance the two terms and is empirically set to 0.1.

**3.4 Face parsing module**

Unlike natural images, FIs are highly structured. That is, each facial component has a statistical pixel intensity distribution, and roughly localizing the facial components is possible. Even for LRHR-FIs with severe degradation, facial components, such as the eyes, nose, and lips, have similar priors. Therefore, face parsing is required to divide input face images into semantic regions to boost the discriminative power of facial features and enhance facial structure expressions.

In the proposed architecture, U-net [39] is chosen as the face parsing module, as shown in Fig. 3.

$$\hat{\mathbf{F}} = f_{\boldsymbol{\theta}_f}(\mathbf{I}, \hat{\mathbf{J}}) \tag{21}$$

Here, $f_{\boldsymbol{\theta}_f}$ indicates the learnable function of the face-parsing module. In Eq. (21), the proposed face parsing module takes two input images: the input LRHR-FI ($\mathbf{I}$) and restored LR-FI ($\hat{\mathbf{J}}$). These two input images are still of low resolution. In the network, as shown in Fig. 3, a concatenation layer is used to stack the LRHR-FI and restored LR-FI. The output of the face-parsing module is the parsed FI. MSE is used as the loss function to learn the face parsing module.

$$L_F = f_{MSE}(\mathbf{F}, \hat{\mathbf{F}}) \tag{22}$$

Here, $L_F$ is the loss used to evaluate the dissimilarity between the original parsed FI ($\mathbf{F}$) and estimated parsed FI ($\hat{\mathbf{F}}$). Figure 5 shows an example of a pair of original HR-FI and its parsed HR-FI. For reference, the CelebA-HQ dataset provides original HR-FIs and their parsed maps.

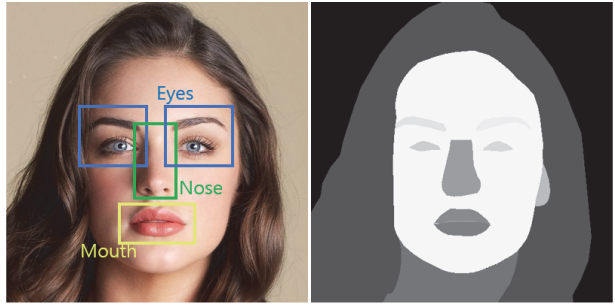

Fig. 6. Original HR-FI (left) and its parsed FI (right)

**3.5 FCG-GAN module**

For HR-FI reconstruction, a new FCG-GAN module is designed. In Fig. 3, the proposed FCG-GAN follows the SRGAN framework, which is one of the basis models for SR. Although recently introduced SR frameworks [28] can be considered, they are too heavy and complex. If the SRGAN is replaced with an advanced SR model, the performance of FCG-GAN improves naturally. The proposed FCG-GAN exhibits differences compared to SRGAN. First, the proposed FCG-GAN employs the parsed LR-FI to increase the discriminative power of facial features and equip the attention mechanism. In FCG-GAN, the parsed LR-FI enables the generator to focus on facial regions and assign different importances to them during training. To this end, a simple approach based on the concatenation layer is used to fuse the parsed LR-FI and feature map output from the generator. More complicated attentions [40] may be considered; however, the concatenation layer is sufficient for a satisfactory performance. In the generator, the concatenation layer is inserted before the first pixel shuffler layer.

$$\hat{\mathbf{H}} = G_\theta(\hat{\mathbf{J}}, \hat{\mathbf{F}}, \mathbf{I}) \tag{23}$$

Here, $G_\theta$ indicates the generator of FCG-GAN. Unlike SRGAN, the proposed generator requires three images: de-rained LR-FI ($\hat{\mathbf{J}}$), parsed LR-FI ($\hat{\mathbf{F}}$), and input LRHR-FI ($\mathbf{I}$).

Second, the proposed FCG-GAN builds local discriminators. Shown in Fig. 3, the global discriminator determines whether the SR-FI is real, whereas the local discriminators determine the authenticity of the facial components of the SR-FI. The details of the facial components are crucial for facial SR. Local discriminators can reinforce the authenticity of facial components, thereby leading to an improvement in facial structure expressions.

$$D_\theta = \{D_{G_\theta}(\widehat{\mathbf{H}}),\ D_{L1_\theta}(cr_e(\widehat{\mathbf{H}})),\ D_{L2_\theta}(cr_n(\widehat{\mathbf{H}})),\ D_{L3_\theta}(cr_l(\widehat{\mathbf{H}}))\} \qquad (24)$$

where $D_{G_\theta}$ and $D_{L_\theta}$ denote global and local discriminators, respectively. In this study, three local discriminators are used for eye, nose, and lip authenticities. In Eq. (24), $cr$ denotes the patch cropping for the eye, nose, and lip regions. For simplicity, the average bounding box positions of the eye, nose, and lip regions are used for patch cropping.

**3.6 Network learning**

Before training the network, each of the heavy-rain removal and face parsing modules is first pretrained. This helps the network super-resolve the LRHR-FI because these modules provide the initial restored LR-FI and parsed LR-FI. Subsequently, the following loss function is minimized to learn the network:

$$L_{G,T} = L_S + \gamma_p L_P + L_G \qquad (25)$$

$$L_S = f_{MSE}(\mathbf{H}, \widehat{\mathbf{H}}) \qquad (26)$$

$$L_P = vgg(\mathbf{H}, \widehat{\mathbf{H}}) \qquad (27)$$

$$L_G = \gamma_1\left(1 - D_\theta\big(G_\theta(\mathbf{I}, \widehat{\mathbf{J}}, \widehat{\mathbf{F}})\big)\right) + \gamma_2\left(1 - D_{L1_\theta}\left(cr_e\big(G_\theta(\mathbf{I}, \widehat{\mathbf{J}}, \widehat{\mathbf{F}})\big)\right)\right) + \gamma_3\left(1 - D_{L2_\theta}(cr_n(G_\theta(\mathbf{I}, \widehat{\mathbf{J}}, \widehat{\mathbf{F}})))\right) +$$
$$\gamma_4\left(1 - D_{L3_\theta}(cr_l(G_\theta(\mathbf{I}, \widehat{\mathbf{J}}, \widehat{\mathbf{F}})))\right) \qquad (28)$$

$$L_D = 1 - D_\theta(\mathbf{H}) + D_\theta(\widehat{\mathbf{H}}) \tag{29}$$

$$L_{D,e} = 1 - D_{L1_\theta}(cr_e(\mathbf{H})) + D_{L1_\theta}(cr_e(\widehat{\mathbf{H}})) \tag{30}$$

$$L_{D,n} = 1 - D_{L2_\theta}(cr_n(\mathbf{H})) + D_{L2_\theta}(cr_n(\widehat{\mathbf{H}})) \tag{31}$$

$$L_{D,l} = 1 - D_{L3_\theta}(cr_l(\mathbf{H})) + D_{L3_\theta}(cr_l(\widehat{\mathbf{H}})) \tag{32}$$

where $L_S$ is the data fidelity term used to measure the dissimilarity between the original HR-FIs $\mathbf{H}$ and restored SR-FIs $\widehat{\mathbf{H}}$., $L_p$ the perceptual loss, and $L_G$ and $L_D$, are related to adversarial loss. Conventional adversarial loss [41] is defined only by the global discriminator $D_\theta$ and generator $G_\theta$, whereas the proposed FCG-GAN adds three loss functions, $D_{L1_\theta}$, $D_{L2_\theta}$, and $D_{L3_\theta}$, to reinforce the authenticity of facial components, as shown in Eqs. (28) and (30)-(32). To train the generator and four discriminators alternatively, $L_{G,T}$ is first minimized with respect to the parameters of the generator. Subsequently, $L_D$, $L_{D,e}$, $L_{D,n}$, and $L_{D,l}$ are minimized with respect to the parameters of each discriminator. That is, the generator and discriminators are updated alternatively in an adversarial relationship with each other. Empirically, $\gamma_p$ and $\gamma_1$ are set to $10^{-3}$ and $\gamma_2, \gamma_3, \gamma_4$ are set to $10^{-4}$. For reference, during network training, heavy-rain removal and face parsing modules use only the MSE for their loss functions.

**3.7 Implementation details**

The CelebA-HQ [34] dataset, which contains celebrity HR-FIs, is used for HR-FI restoration. A total of 18,000, 1,800, and 100 datasets are used as training, validation, and test datasets, respectively, to train and test the network. The image size of the HR-FIs in the CelebA-HQ dataset is 128 × 128 pixels. To effectively train the network, the heavy rain and face parsing modules are pretrained with the training dataset, as mentioned in the previous section. The batch size is 64, and the number of epochs is 200. The Adam [42] optimizer is used with a weight decay $10^{-4}$, and the learning rate is set to $10^{-3}$. The scale factor of *s* in Eq. (5) is set to 4, and a bicubic interpolation is used to generate LR-FIs. In addition, *m* is set to one such that one rain streak is added. For network training, the Adam optimizer is also applied with the same parameter settings, except that the epoch is 80 and batch size is 16. The proposed network is implemented using the PyTorch framework. Our source codes and datasets will be available from https://github.com/cvmllab.

## 4. Results

This section evaluates the proposed method using a few baseline methods for our heavy-rain face image dataset. First, the SRGAN model is compared with the proposed heavy rain face image restoration. This comparison shows the effectiveness of the proposed unified framework in improving visibility and resolution simultaneously. In addition, an ablation study verifies that facial structure expressions can be improved step-by-step by adding face parsing and FCG-GAN. Second, SOTA methods such as heavy-rain removal [8], image-to-image translation [41], SRGAN [24], face SR [29], and ESRGAN [27] are compared. This evaluation confirms that the proposed unified framework is more suitable than the SOTA method for joint heavy-rain removal and SR. The proposed method is shown to outperform SOTA methods.

### 4.1. Ablation study

Figure 7 shows the results for the heavy-rain face image restoration. The first and second columns show the input LRHR-FIs and corresponding HR-FIs, respectively. According to the proposed scale-aware heavy rain model, LRHR-FIs were generated from the original HR-FIs. As shown in the first column, LRHR-FIs were significantly degraded in terms of both visibility and resolution. In addition, rain streaks were added. Thus, solving the inverse problem of the scale-aware heavy-rain model is challenging.

An ablation study is required to verify the effectiveness of the three proposed modules: the heavy rain removal module (HRRM), face parsing module (FPM), and FCG-GAN. As mentioned in Section 3.5, the SRGAN framework was adopted for the SR module in the proposed network. Therefore, the proposed FCG-GAN can be considered an advanced SRGAN. The third column of Fig. 7 provides the SR-FIs restored using SRGAN [24]. The fourth and fifth columns show the SR-FIs reconstructed using the proposed methods: HRRM+ FPM+SRGAN and HRRM+FPM+FCG-GAN. The difference between the two proposed methods is the application of the local discriminators.

First, by comparing the SR-FIs in the third and fourth columns, the use of HRRM and FPM is known to significantly improve face resolution. In particular, facial components such as the eyes, nose, and lips were more clearly restored using the proposed method (HRRM+ FPM+SRGAN). The parsed LR-FIs enable the generator to focus on facial regions and provide them different levels of importance. The parsed LF-FIs serve as a guide to teach the generator, whose features are more informative, thereby increasing the discriminative power of facial features and improving the facial structure expression. Moreover, when only the SRGAN is applied, the facial components are blurred and distorted, as shown in third column. Because SRGAN was originally designed for SR, it is unsuitable for super-resolving LRHR-FIs. That is, the HRRM was excluded in the SRGAN.

Second, the additional use of local discriminators can further enhance facial structure expressions. The fifth column shows the SR-FIs reconstructed using the proposed method (HRRM+FPM+FCG-GAN). Compared with the SR-FIs in the fourth column, the shapes of the facial components are relatively more refined and closer to the ground truth HR-FIs, as shown in the red boxes. Note that the sharpness around the eyes and nose is further enhanced. This indicates that the addition of local discriminators reinforces the authenticity of facial components and improves facial structure expression.

**4.2. Visual quality evaluation**

The proposed method (HRRM+FPM+FCG-GAN) is compared with SOTA methods such as SRGAN [24], heavy-rain removal [8], pix2pix [41], face SR [29], and ESRGAN [27] for visual quality comparison. The SRGAN does not contain the HRRM; thus, the visual quality of the reconstructed images is poor, as shown in third column. In particular, facial components such as the eyes and lips are blurred.

The sixth and seventh columns show the SR-FIs reconstructed using the heavy-rain removal and pix2pix methods, respectively. In these columns, severe distortions appeared around the facial components. In the pix2pix method [41], neither the IDM for heavy rain nor SR model was reflected. Thus, the CGAN used in pix2pix fails to generate a face image intensity distribution. The heavy-rain removal method [8] was originally designed for rain removal, requiring the consideration of a physics-based network. This implies that the model architecture is unsuitable for SR. Consequently, the SRGAN performed better than the heavy-rain removal method. For reference, to directly apply the pix2pix and heavy-rain removal methods to the heavy-rain face image restoration, the size of the LRHR-FIs was changed to the same size as the original HR-FIs.

The last column shows the SR-FIs reconstructed using the face SR method. In this column, the resolution improved more than that of the SRGAN, owing to the use of parsed LR-FIs. In this study, to implement the face SR method [29], the parsed LR-FI and LRHR-FI were fed to the input layer of the SRGAN. For the pair comparison, the same SRGAN model was used in the face SR method. That is, the difference between the SRGAN and face SR method is whether a parsed LR-FI is used.

However, the face SR method is inferior to the proposed method (HRRM+FPM+FCG-GAN). In particular, better sharpness was achieved using the proposed method. For example, the first row confirms that the proposed method produces a clearer pair of eyes and nose than the facial SR method. In addition, the shape of the eyes and nose, or face outline, was closer to the original HR-FIs. This indicates that the use of local discriminators is effective in improving facial structure expressions. In the proposed FCG-GAN, local discriminators reinforce the authenticity of facial components, thereby inducing the generator to produce more realistic images. Consequently, the proposed method can restore SR-FIs with sharper and more accurate shapes than the face SR method.

Fig. 8 shows a comparison of the proposed method with the ESRGAN method. Although ESRGAN is a SOTA method for SR, the visual quality is evidently poor because of geometric distortion around facial components. Overall, it was blurry. This is because the model architecture is designed for SR, and thus, solving the inverse problem of the scale-aware heavy-rain model is limited.

Table 1. Performance evaluation.

| Methods / Metrics | PSNR | SSIM |
|---|---|---|
| Heavy-Rain Removal [8] | 20.8678 | 0.5640 |
| Pix2Pix [41] | 21.8714 | 0.6052 |
| SRGAN [24] | 22.4878 | 0.6852 |
| Face SR [29] | 22.7130 | 0.6988 |
| ESRGAN [27] | 23.0400 | 0.6977 |
| Proposed Method (HRRM+ FPM+SRGAN) | 22.7340 | 0.6981 |
| **Proposed Method (HRRM+FPM+FCG-GAN)** | **23.2075** | **0.7120** |

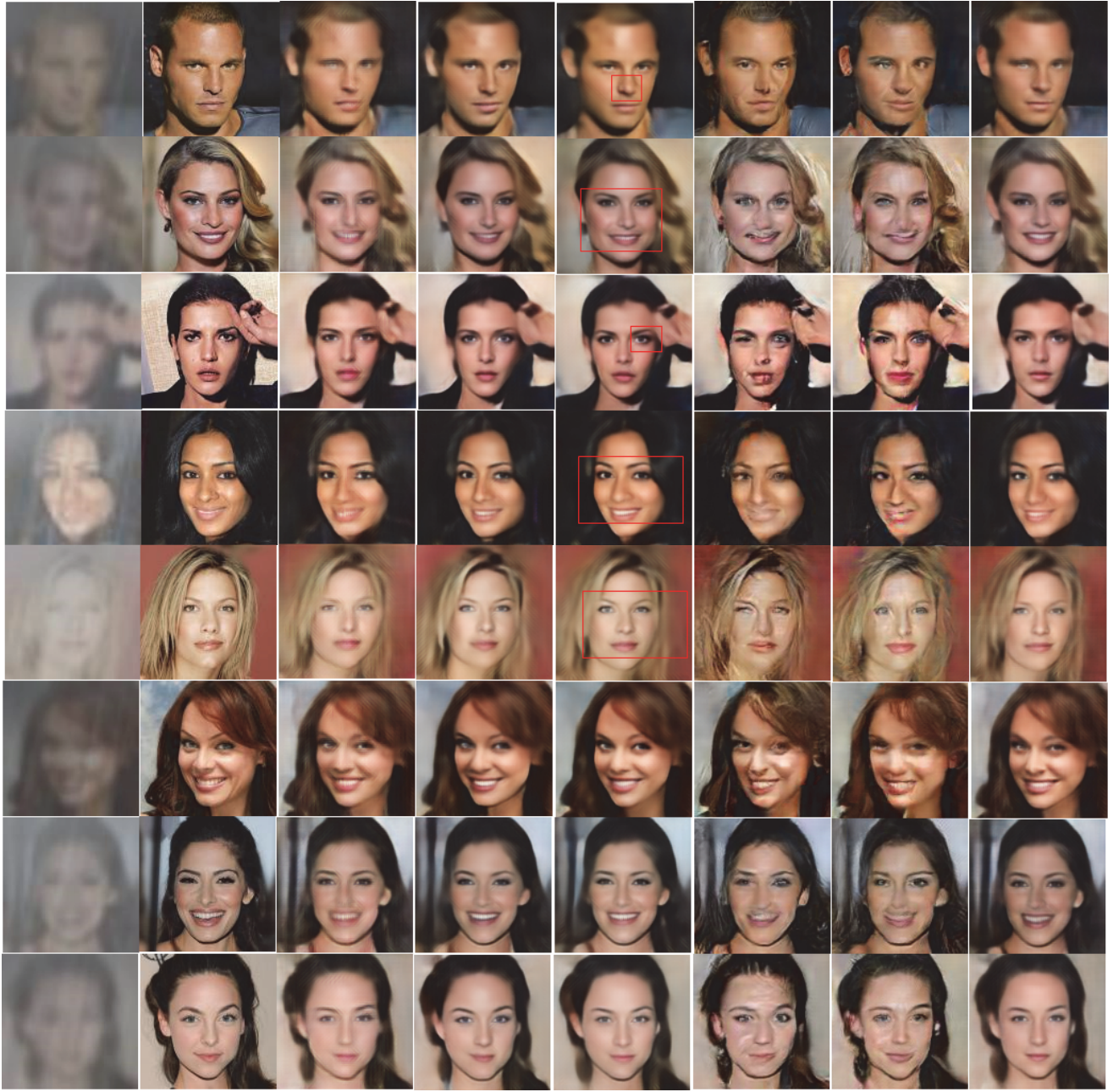

Fig. 7. Results: input LRHR-FIs, ground truth HR-FIs, SRGAN [24], proposed method (HRRM+FPM+SRGAN), proposed method (HRRM+FPM+FCG-GAN), heavy-rain removal [8], pix2pix [41], and face SR [29] (left to right).

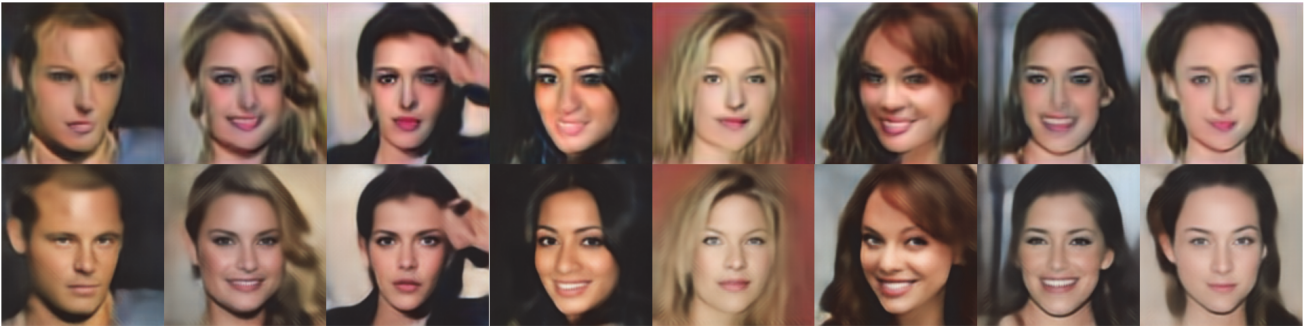

Fig. 8. Results: ESRGAN [27] (top row) and proposed method (bottom row).

**4.3. Quantitative evaluation**

For performance evaluation, the peak signal-to-noise ratio (PSNR) and structural similarity (SSIM) [43] were tested. The PSNR measures the sum of the pixel-wise differences between two images in a log space to reflect the human visual system, and the SSIM scores the structural similarity between two images based on the luminance, standard deviation, and contrast functions. A higher value indicates a higher quality for both PSNR and SSIM. Although other no-reference quality metrics are available [44], this study aims to reconstruct SR-FIs with resolutions and visibilities as similar to the original HR-FIs as possible. Therefore, such no-reference quality metrics may be inaccurate, and thus, inappropriate for this study.

Table 1 lists the results of the PSNR and SSIM evaluations for the test dataset. As expected, the proposed method (HRRM+FPM+FCG-GAN) demonstrated the best performance among all methods and surpassed the SOTA methods. This indicates that the proposed FCGAL is effective in improving facial structure expressions. That is, FPM serves as a guide to teach FCG-GAN, whose features are more informative, thereby increasing the discriminative power of facial features and improving facial structure expression. In addition, local discriminators can reinforce the authenticity of facial components, which induces the generator to produce more realistic and clearer SR-FIs. Unlike conventional SOTA methods, the proposed method integrates the heavy rain and resolution conversion models to address joint heavy-rain removal and SR. Therefore, the proposed physics-based network is more suitable for super-resolving LRHR-FIs than the SOTA methods.

From the results, the proposed method conclusively outperforms the SOTA methods. Conventional pix2pix,

heavy-rain removal, and SR methods are unsuitable for joint heavy-rain removal and SR. By contrast, the proposed network incorporates heavy rain and resolution conversion models to solve the inverse problem of the scale-aware heavy rain model. For heavy-rain removal, an interpretable IDM-based network was designed for physical parameter estimation, and the proposed FCGAL was applied based on a face-parsing-guided generator and local discriminators to handle facial features and reinforce the authenticity of facial components. Based on this approach, heavy-rain can be removed and the image resolution can be increased, whereas conventional SOTA methods yield blurred SR-FIs. Geometric distortion occurs, depending on the method used. Moreover, the shape of facial components is incorrect. This is because their architectures were originally designed for a single task, such as either heavy-rain removal or SR.

### 4.4. Discussion

The proposed method was only applied to synthetic LRHR-FIs. Shooting real LRHR-FIs is forbidden because of personal information. In addition, everyone in our country is wearing masks because of COVID-19. Therefore, collecting real LRHR-FIs is difficult. Moreover, real LRHR-FIs can have different appearances from the synthesized images using the proposed scale-aware heavy rain model. Thus, more complex IDM is required. For example, the model parameters for downsampling and Gaussian blurring are not fixed or viable. A more complex blind heavy-rain face image restoration approach [33] is required for super-resolving real LRHR-FIs. In future work, we plan to upgrade the proposed network for blind heavy-rain face image restoration such that the proposed method can be more applicable to real-world scenarios.

### 5. Conclusion

This paper presents a learning method for restoring heavy-rain face images with severely low visibility and resolution. Unlike conventional SR and heavy-rain removal, a scale-aware heavy rain model that integrates IDMs for heavy rain and low-resolution conversion is introduced. To effectively solve the inverse problem of the proposed scale-aware heavy rain model, the proposed network was constructed based on three modules:

HRRM, FPM, and FCG-GAN. The HRRM is used for heavy-rain removal and consists of one encoder and four decoders to extract visual features and predict physical parameters. The HRRM can be viewed as a simple interpretable IDM-based network. The FPM takes LR-FIs and LRHR-FIs as input and generates parsed LR-FIs to identify facial areas. The output map of the FPM is inserted into the FCG-GAN to teach which features are informative for face SR. This increases the discriminative power of facial features and leads to an improvement in image resolution. In addition, the use of local discriminators can reinforce the authenticity of facial components, thereby improving facial structure expressions. The proposed network can remove heavy rain and simultaneously increase resolution and visibility. The results confirmed that the proposed model surpasses the SOTA methods and is more suitable for joint heavy-rain removal and SR than conventional methods such as image-to-image translation, heavy-rain removal, and SR models.

**Acknowledgment**

This work was supported by the National Research Foundation of Korea(NRF) grant funded by the Korea government(MSIT) (No. 2020R1A2C1010405).